\documentclass[conference]{IEEEtran}
\IEEEoverridecommandlockouts
\usepackage{cite}
\usepackage{amsmath,amssymb,amsfonts}
\usepackage{algorithmic}
\usepackage{graphicx}
\usepackage{textcomp}
\usepackage{xcolor}
\usepackage{booktabs}
\def\BibTeX{{\rm B\kern-.05em{\sc i\kern-.025em b}\kern-.08em
    T\kern-.1667em\lower.7ex\hbox{E}\kern-.125emX}}
\makeatletter
 \let\old@ps@headings\ps@headings
 \let\old@ps@IEEEtitlepagestyle\ps@IEEEtitlepagestyle
 \def\confheader#1{%
 \def\ps@IEEEtitlepagestyle{%
 \old@ps@IEEEtitlepagestyle%
 \def\@oddhead{\strut\hfill#1\hfill\strut}%
 \def\@evenhead{\strut\hfill#1\hfill\strut}%
 }%
 \ps@headings%
 }
 \makeatother

\confheader{%
 32nd SIGKDD Conference on Knowledge Discovery and Data Mining - AI for Sciences Track
 }

 \usepackage[pscoord]{eso-pic}
\newcommand{\placetextbox}[3]{
 \setbox0=\hbox{#3}
 \AddToShipoutPictureFG*{ \put(\LenToUnit{#1\paperwidth},\LenToUnit{#2\paperheight}){\vtop{{\null}\makebox[0pt][c]{#3}}}
 }
 }
 \placetextbox{.23}{0.055}{\small{979-8-3503-8166-5/24/\$31.00~\copyright 2024 IEEE}}

\begin{document}

\title{AI-Driven Early Warning Systems for Student Success: Discovering Static Feature Dominance in Temporal Prediction Models}

\author{\IEEEauthorblockN{1\textsuperscript{st} Vaarunay Kaushal}
\IEEEauthorblockA{\textit{Data Science and Computer Applications} \\
\textit{Manipal Institute of Technology, MAHE}\\
Manipal, India \\
vaarunay.kaushal@learner.manipal.edu}
\and
\IEEEauthorblockN{2\textsuperscript{nd} Dr. Rajib Mall}
\IEEEauthorblockA{\textit{Computer Science and Engineering} \\
\textit{Shiv Nadar University }\\
Greater Noida, Uttar Pradesh \\
rajib.mall@snu.edu.in}
}

\maketitle

\begin{abstract}
Early identification of at-risk students is critical for effective intervention in online learning environments. This study extends temporal prediction analysis to Week 20 (50\% of course duration), comparing Decision Tree and Long Short-Term Memory (LSTM) models across six temporal snapshots. Our analysis reveals that different performance metrics matter at different intervention stages: high recall is critical for early intervention (Weeks 2-4), while balanced precision-recall is important for mid-course resource allocation (Weeks 8-16), and high precision becomes paramount in later stages (Week 20). We demonstrate that static demographic features dominate predictions (68\% importance), enabling assessment-free early prediction. The LSTM model achieves 97\% recall at Week 2, making it ideal for early intervention, while Decision Tree provides stable balanced performance (78\% accuracy) during mid-course. By Week 20, both models converge to similar recall (68\%), but LSTM achieves higher precision (90\% vs 86\%). Our findings also suggest that model selection should depend on intervention timing, and that early signals (Weeks 2-4) are sufficient for reliable initial prediction using primarily demographic and pre-enrollment information.
\end{abstract}

\section{Introduction}

Timely prediction of at-risk students is a fundamental challenge in online education, with significant implications for student retention and success. Existing methods concentrate on detecting at-risk students with high confidence at time points that are close to the end of the course \cite{b1}.However, it is crucial to identify at-risk students more quickly, so there is sufficient time for intervention and improvement.

This study addresses several critical gaps in the existing literature. First, while most studies in the existing literature focus on prediction closer to course completion, we limit temporal analysis to Week 20, marking 50\% of the course duration. Second, we compare Decision Tree and LSTM models across multiple temporal snapshots, revealing that different models excel at different intervention stages. Third, we demonstrate that static demographic features dominate predictions (68\% importance), enabling assessment-free early prediction. Finally, we establish a model selection framework based on intervention timing and performance metric priorities.

Our key contributions include: (1) extended temporal analysis revealing performance patterns across the first half of courses, (2) assessment-free prediction framework enabling well timed intervention, (3) discovery that static features dominate temporal engagement features, (4) model selection framework matching models to intervention needs, and (5) finding that early signals (Weeks 2-4) are sufficient for reliable initial prediction and can be trusted.

\section{Literature Review}

Several studies have investigated student performance prediction using various machine learning approaches. Adnan et al. \cite{b1} developed several predictive models based on Machine Learning and Deep Learning algorithms, examining various sets of variables such as demographic attributes, clickstream data, and assessment-related information. Nonetheless, their temporal analysis was confined to discrete course completion milestones (20\%, 40\%, 60\%, 80\%, 100\%) rather than continuous weekly intervals, and the most encouraging performance metrics appeared only near the end of the course, where they have little practical value.

Buschetto Macarini et al. \cite{b2} created 13 unique datasets reflecting varying dimensions of student engagement within online learning spaces. B. Minaei-Bidgoli et al. \cite{b3} introduced methods for categorizing students to forecast final grades using logged information. J. Xu et al. \cite{b4} proposed a two-layered framework with foundational and ensemble predictors for forecasting outcomes based on dynamic performance states.

Recent work has explored deep learning approaches. Q. Liu et al. \cite{b5} introduced Exercise-Enhanced Recurrent Neural Network (EERNN) frameworks. Nannan Wu et al. \cite{b9} created CLMS-Net, integrating convolutional, LSTM, and MLP components for MOOC attrition prediction. Mao et al. \cite{b10} implemented time-aware LSTM (T-LSTM) for predicting student success, though noting limitations in generalization.

A critical gap in existing literature is the lack of recognition that time-variable features such as assessment scores may not be as important for prediction as previously assumed. While many studies incorporate assessment data and temporal engagement metrics, few investigate whether these time-dependent features are actually necessary for reliable early prediction. This study addresses this gap by demonstrating that static demographic features dominate predictions, enabling earlier intervention with high confidence using primarily pre-enrollment information, without waiting for assessment performance data to accumulate.

\section{Methodology}

\subsection{Dataset}

We utilize the Open University Learning Analytics Dataset (OULAD) \cite{oulad}, which contains anonymized data from 32,593 students across 7 courses and 22 presentations. The dataset includes student demographics, VLE (Virtual Learning Environment) interactions, assessment submissions, and final outcomes. Course duration ranges from 269-273 days, with our analysis covering the first 140 days (approximately 50\% of course duration).

\subsection{Data Preprocessing}

Data preprocessing involved merging seven OULAD datasets: studentInfo, studentVle, studentAssessment, courses, assessments, studentRegistration, and vle. We created a binary target variable where ``at-risk'' is defined as students who either ``Withdrew'' or ``Failed'' the course, while ``not at-risk'' includes students who ``Passed'' or achieved ``Distinction.''

Baseline data preparation involved aggregating data per student-course combination using \texttt{groupby(['id\_student', 'code\_module', 'code\_presentation'])} with \texttt{sum\_click} summed, resulting in 32,593 unique student-course entries. The baseline Decision Tree model achieves 76.85\% accuracy on this aggregated dataset.

\subsection{Feature Engineering}

We engineered 11 features for model training:
\begin{itemize}
    \item \textbf{Demographics (8 features)}: gender, code\_module, code\_presentation, region, highest\_education, IMD\_band, age\_band, disability
    \item \textbf{Academic History (2 features)}: num\_of\_prev\_attempts, studied\_credits
    \item \textbf{Engagement (1 feature)}: sum\_click (cumulative VLE clicks)
\end{itemize}

Notably, we created assessment features (\texttt{assessments\_submitted\_day} and \texttt{avg\_assessment\_score\_day}) but \textit{did not} include them in the baseline model, enabling assessment-free prediction. This allows intervention before assessment performance is available.

All categorical features were label-encoded, and IMD\_band was binned into 5 categories (VeryLow, Low, Medium, High, VeryHigh).

\subsection{Temporal Snapshot Creation}

We created six temporal snapshots corresponding to Weeks 2, 4, 8, 12, 16, and 20 (Days 14, 28, 56, 84, 112, 140) for testing model performance across these snapshots. For each snapshot, we filtered data to include only interactions occurring up to that day, then aggregated features per student-course combination.

The rationale for analyzing 50\% of course duration is that intervention timing is critical—earlier intervention is more effective and leaves time for implementing, gauging and calibrating strategies, making the first half of the course the most relevant period for predictive modeling.

\subsection{Model Selection}

We compare two model architectures:

\textbf{Decision Tree}: A baseline interpretable model trained on full-course data and tested on temporal snapshots without retraining. This prevents temporal leakage and ensures fair comparison across snapshots.

\textbf{LSTM}: A deep learning model with two LSTM layers (64→32 units), dropout (0.3-0.4), and class weights for handling imbalance. The model processes sequences of 84 timesteps, with each timestep containing 11 features (10 static + 1 temporal). For shorter snapshots (e.g., Week 2 with 14 days), sequences are padded with zeros to 84 timesteps.

\subsection{Evaluation Strategy}

We evaluate models using accuracy, precision, recall, F1-score, and ROC-AUC. To prevent temporal leakage, we use a single model (trained on full data) tested on temporal snapshots, rather than training separate models per snapshot. This ensures that predictions at a week use only information available at that time, not future information.

\section{Results}

\subsection{Baseline Performance}

The Decision Tree baseline model achieves 76.85\% accuracy, 78.04\% precision, 78.15\% recall, and 78.10\% F1-score. Feature importance analysis reveals that demographic features dominate predictions, accounting for 68\% of total importance, with region (27.66\%), studied\_credits (13.50\%), and IMD\_band (11.39\%) being the top three features. Temporal engagement (sum\_click) has minimal importance (1.34\%).

\begin{table}[!h]
\centering
\caption{Baseline Decision Tree Performance}
\label{tab:baseline}
\begin{tabular}{lcc}
\toprule
\textbf{Metric} & \textbf{Value} & \textbf{Percentage} \\
\midrule
Accuracy & 0.7685 & 76.85\% \\
Precision & 0.7804 & 78.04\% \\
Recall & 0.7815 & 78.15\% \\
F1-Score & 0.7810 & 78.10\% \\
ROC-AUC & 0.7678 & 76.78\% \\
\bottomrule
\end{tabular}
\end{table}

\subsection{Temporal Performance - Decision Tree}

Table \ref{tab:dt_temporal} shows Decision Tree performance across temporal snapshots. Performance remains remarkably stable, with accuracy ranging from 75.88\% (Week 2) to 78.03\% (Week 8). A clear precision-recall trade-off emerges: precision increases from 74.23\% (Week 2) to 85.93\% (Week 20), while recall decreases from 83.22\% (Week 2) to 68.17\% (Week 20). This suggests the model becomes more conservative as more data accumulates, prioritizing precision over recall.

\begin{table}[!h]
\centering
\caption{Decision Tree Temporal Performance}
\label{tab:dt_temporal}
\begin{tabular}{lcccc}
\toprule
\textbf{Week} & \textbf{Accuracy} & \textbf{Precision} & \textbf{Recall} & \textbf{F1-Score} \\
\midrule
Week 2 & 0.7588 & 0.7423 & 0.8322 & 0.7847 \\
Week 4 & 0.7791 & 0.8066 & 0.7650 & 0.7852 \\
Week 8 & 0.7803 & 0.8443 & 0.7159 & 0.7748 \\
Week 12 & 0.7772 & 0.8533 & 0.6980 & 0.7679 \\
Week 16 & 0.7768 & 0.8639 & 0.6853 & 0.7643 \\
Week 20 & 0.7730 & 0.8593 & 0.6817 & 0.7602 \\
\bottomrule
\end{tabular}
\end{table}

\begin{figure}[!h]
\centering
\includegraphics[width=0.45\textwidth]{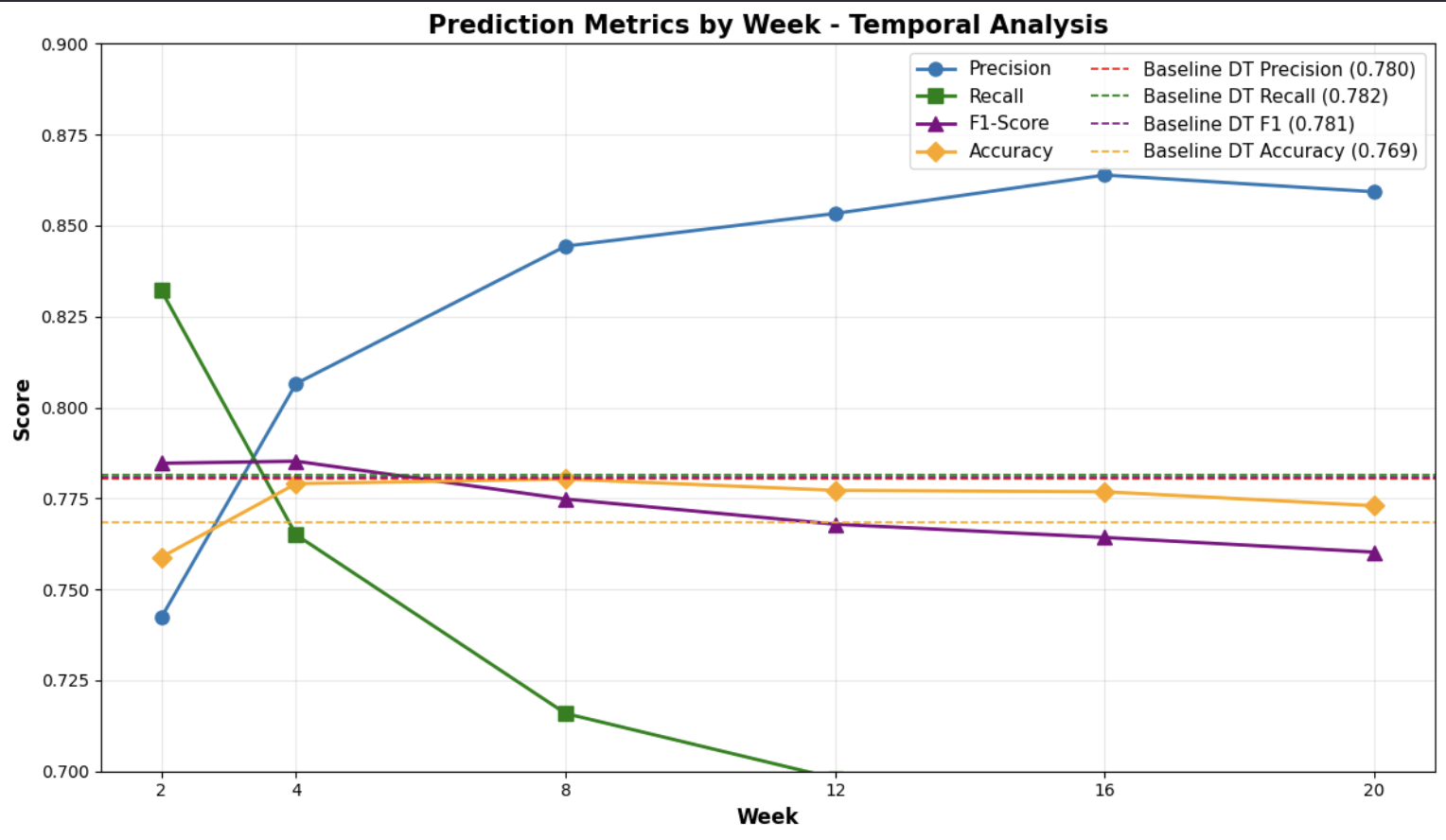}
\caption{Decision Tree performance metrics across temporal snapshots (Weeks 2-20). Performance remains stable with accuracy 75.88\%-78.03\%, while precision increases (74.23\% → 85.93\%) and recall decreases (83.22\% → 68.17\%) over time.}
\label{fig:dt_temporal}
\end{figure}

\subsection{Temporal Performance - LSTM}

Table \ref{tab:lstm_temporal} shows LSTM performance across temporal snapshots. Unlike Decision Tree, LSTM demonstrates dramatic improvement over time: accuracy increases from 53.80\% (Week 2) to 80.00\% (Week 20). This improvement correlates with reduced sequence padding—Week 2 sequences are heavily padded (14 days of data + 70 days of zeros), while Week 20 sequences use truncation (last 84 days of 140-day data).

LSTM shows extremely high recall early (97.31\% at Week 2, 93.02\% at Week 4), making it ideal for early intervention when missing at-risk students is costly. By Week 20, LSTM achieves high precision (90.00\%) while maintaining the same recall as Decision Tree (68\%).

\begin{table}[!h]
\centering
\caption{LSTM Temporal Performance}
\label{tab:lstm_temporal}
\begin{tabular}{lcccc}
\toprule
\textbf{Week} & \textbf{Accuracy} & \textbf{Precision} & \textbf{Recall} & \textbf{F1-Score} \\
\midrule
Week 2 & 0.5380 & 0.5336 & 0.9731 & 0.6942 \\
Week 4 & 0.6363 & 0.6018 & 0.9302 & 0.7308 \\
Week 8 & 0.7000 & 0.6900 & 0.7900 & 0.7400 \\
Week 12 & 0.7500 & 0.7600 & 0.7500 & 0.7600 \\
Week 16 & 0.7400 & 0.7600 & 0.7300 & 0.7400 \\
Week 20 & 0.8000 & 0.9000 & 0.6800 & 0.7800 \\
\bottomrule
\end{tabular}
\end{table}

\begin{figure}[!h]
\centering
\includegraphics[width=0.45\textwidth]{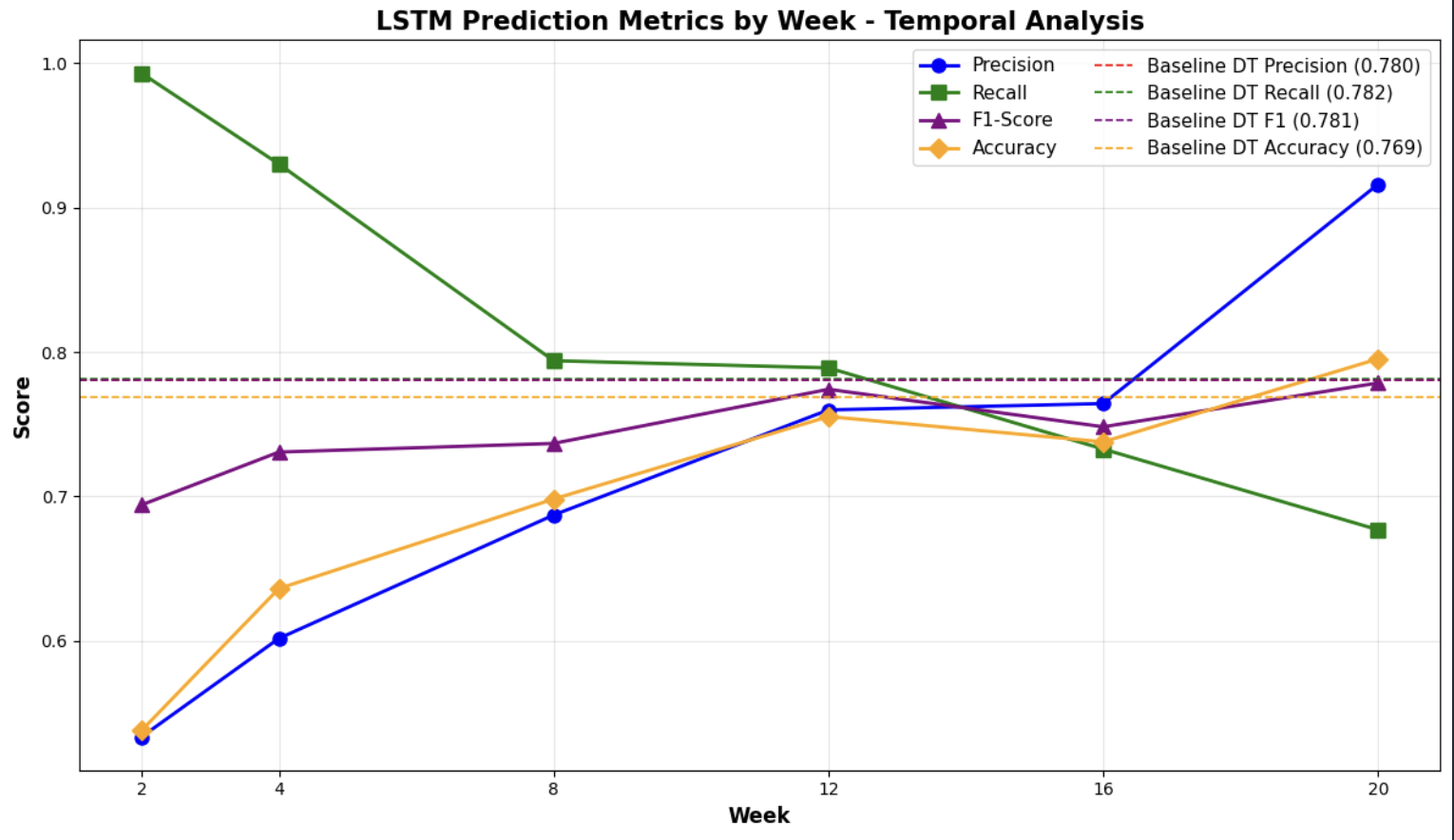}
\caption{LSTM performance metrics across temporal snapshots (Weeks 2-20). Accuracy improves dramatically from 53.80\% (Week 2) to 80.00\% (Week 20), with extremely high recall early (97.31\% at Week 2) and high precision late (90.00\% at Week 20).}
\label{fig:lstm_temporal}
\end{figure}

\subsection{Model Comparison}

Figure \ref{fig:comparison} compares LSTM and Decision Tree across all metrics. Key insights emerge:

\begin{itemize}
\item \textbf{Early Weeks (2-4)}: LSTM achieves significantly higher recall (97.31\% vs 83.22\% at Week 2, 93.02\% vs 76.50\% at Week 4) compared to Decision Tree, though with lower precision (53.36\% vs 74.23\% at Week 2).

\item \textbf{Mid-Course (8-16)}: Decision Tree provides stable balanced performance (78\% accuracy, 84-86\% precision, ~72\% recall), while LSTM shows greater variability (70-75\% accuracy) during this period.

\item \textbf{Late Course (Week 20)}: LSTM achieves higher precision (90\% vs 86\%) with the same recall (68\%) as Decision Tree.
\end{itemize}

\begin{figure}[!h]
\centering
\includegraphics[width=0.45\textwidth]{images/LSTMvDecision_tree.png}
\caption{Comparison of LSTM and Decision Tree performance across temporal snapshots. Top-left: Accuracy comparison showing LSTM's dramatic improvement (54\% → 80\%) vs Decision Tree's stability (76\%-78\%). Top-right: Precision comparison showing LSTM's improvement (53\% → 90\%) vs Decision Tree's steady increase (74\% → 86\%). Bottom-left: Recall comparison showing LSTM's high early recall (97\% → 68\%) vs Decision Tree's steady decline (83\% → 68\%). Bottom-right: F1-Score comparison showing convergence by Week 20.}
\label{fig:comparison}
\end{figure}

\subsection{Feature Importance Analysis}

Feature importance analysis (Table \ref{tab:features}) reveals that static demographic features dominate predictions, accounting for 68\% of total importance. Region (27.66\%) is the most important feature, followed by studied\_credits (13.50\%) and IMD\_band (11.39\%). Temporal engagement (sum\_click) has minimal importance (1.34\%), despite increasing 370\% from Week 2 to Week 20.

This finding has profound implications: early prediction is feasible using primarily pre-enrollment demographic information, with minimal benefit from temporal engagement data. Assessment features, while created, are not used, enabling assessment-free prediction.

\begin{table}[!h]
\centering
\caption{Top 8 Feature Importances (Decision Tree Baseline)}
\label{tab:features}
\begin{tabular}{lc}
\toprule
\textbf{Feature} & \textbf{Importance} \\
\midrule
Region & 0.2766 (27.66\%) \\
Studied Credits & 0.1350 (13.50\%) \\
IMD Band & 0.1139 (11.39\%) \\
Code Presentation & 0.0958 (9.58\%) \\
Code Module & 0.0948 (9.48\%) \\
Age Band & 0.0773 (7.73\%) \\
Gender & 0.0723 (7.23\%) \\
Highest Education & 0.0620 (6.20\%) \\
\midrule
\textbf{Demographics Total} & \textbf{0.6817 (68.17\%)} \\
\textbf{Engagement (Clicks)} & \textbf{0.0134 (1.34\%)} \\
\bottomrule
\end{tabular}
\end{table}


\section{Discussion}

\subsection{Early Intervention Needs (Weeks 2-4)}

Early intervention is most effective when implemented promptly, making high recall critical during Weeks 2-4. The cost of missing an at-risk student (false negative) far exceeds the cost of incorrectly flagging a non-at-risk student (false positive) at this stage. Our results show that LSTM achieves 97.31\% recall at Week 2, compared to 83.22\% for Decision Tree—a 14 percentage point advantage that translates to catching 14\% more at-risk students.

Assessment-free prediction enables Week 2 intervention, as our model uses only demographic information and early engagement patterns (cumulative clicks up to Day 14). This is 2-6 weeks earlier than assessment-based approaches that require first assessment submissions (typically Week 4-8).

\subsection{Mid-Course Balanced Performance (Weeks 8-16)}

During mid-course (Weeks 8-16), intervention resources are being strategically allocated, making both precision and recall important. Decision Tree provides superior balanced performance during this period: stable 78\% accuracy, 84-86\% precision, and consistent F1-scores (0.76-0.77). This stability is crucial for premature resource allocation plans, where predictability matters more than peak performance.

Additionally, Decision Tree's interpretability advantage becomes important during mid-course. Educators need to understand why students are flagged to design targeted interventions. Decision Tree's feature importance and decision rules provide this transparency, building trust in the system.

\subsection{Late-Course High Precision (Week 20)}

By Week 20, educators have accumulated sufficient information through direct observation, assessment performance, and engagement patterns to identify at-risk students. At this stage, the model's priority shifts to flagging students with high confidence, accepting that it will miss approximately 32\% of at-risk students (68\% recall). This trade-off is acceptable because educators already have their own understanding of the at-risk student pool through direct observation and accumulated assessment data.

LSTM achieves 90\% precision at Week 20, compared to 86\% for Decision Tree, while both models maintain 68\% recall. The LSTM's precision advantage makes it preferable for late-course prediction. Importantly, any at-risk students missed by the model but present in the educator's pool can be verified by the educators themselves. The model's high-confidence predictions serve to reinforce the common at-risk pool between educators and the model, providing additional validation and confidence in intervention decisions rather than replacing educator judgment.

\subsection{Static Feature Dominance}

Our analysis reveals that static demographic features (68\% importance) dominate temporal engagement features (1.34\% importance). This finding challenges assumptions about engagement-based early warning systems and has several implications:

\begin{itemize}
    \item \textbf{Early Prediction Feasibility}: Pre-enrollment demographic information is sufficient for reliable early prediction, with minimal benefit from temporal engagement data.
    \item \textbf{Assessment-Free Prediction}: Assessment scores are not required for early prediction, enabling Week 2 intervention.
    \item \textbf{Structural Factors}: Geographic region (27.66\%) and socioeconomic status (IMD band: 11.39\%) are also strong predictors, highlighting structural inequities in educational outcomes.
    \item \textbf{Intervention Timing}: Early signals (Weeks 2-4) contain most predictive information, making early intervention both feasible and critical.
\end{itemize}

\subsection{Why 50\% of Course Duration}

Our analysis extends to Week 20 (50\% of course duration, ~140 days of ~269 days) for several reasons:

\begin{itemize}
    \item \textbf{Actionability}: At the Week 20 mark, educators have sufficient information through direct observation and accumulated assessment data to identify at-risk students. Beyond this timeframe, predictions becomes less actionable.
    \item \textbf{Intervention Timing}: Early intervention is more effective than late intervention. The first half of the course represents the critical window for predictive modeling.
    \item \textbf{Resource Allocation}: Mid-course (Weeks 8-16) is when intervention resources are strategically allocated in advance, making this period crucial for prediction.
    \item \textbf{Practical Considerations}: Analyzing beyond 50\% would require predicting outcomes that are already evident, reducing the practical value of the prediction system.
\end{itemize}


\subsection{Model Selection Philosophy}

Our results demonstrate that model selection should depend on intervention timing and performance metric priorities. We propose the following framework:

\begin{itemize}
    \item \textbf{Early Intervention (Weeks 2-4)}: Prioritize recall → Use LSTM (97\% recall at Week 2)
    \item \textbf{Mid-Course Resource Allocation (Weeks 8-16)}: Prioritize balanced performance and interpretability → Use Decision Tree (78\% accuracy, interpretable)
    \item \textbf{Late-Course Precision (Week 20)}: Prioritize precision → Use LSTM (90\% precision)
\end{itemize}

This framework recognizes that different intervention stages have different requirements, and model selection should match these needs rather than seeking a single ``best'' model.


\section{Limitations and Future Work}

Several limitations should be acknowledged. First, assessment features were created but not included in the baseline model. Future work should explore incorporating assessment data to improve mid-to-late course prediction. Second, we used a single Decision Tree architecture; comparing with Random Forest, XGBoost, or ensemble methods could reveal additional insights. Third, the LSTM model uses mostly static features, which is computationally inefficient. Future work should explore LSTM architectures with temporal-only features or hybrid approaches combining static and temporal features.

The performance(f1-score) decline after Week 8 for Decision Tree needs investigation. Possible causes include model architecture mismatch, feature distribution shifts, or overfitting to full-course patterns. Future work should explore snapshot-specific models or adaptive learning approaches.

Finally, fairness analysis was not included in this study. Given that demographic features dominate predictions, analyzing potential biases across demographic groups is crucial for ethical deployment. Future work should include comprehensive fairness analysis and bias mitigation strategies.

\section{Conclusion}

This study extends temporal prediction analysis to Week 20 (50\% of course duration), comparing Decision Tree and LSTM models across six temporal snapshots. Our key findings include:

\begin{enumerate}
    \item \textbf{Extended Temporal Analysis}: Performance patterns reveal that different metrics matter at different intervention stages, with LSTM excelling early (high recall) and late (high precision), while Decision Tree provides stable balanced performance mid-course.
    \item \textbf{Assessment-Free Prediction}: Static demographic features (68\% importance) enable early prediction without assessment data, allowing Week 2 intervention—2-6 weeks earlier than assessment-based approaches.
    \item \textbf{Model Selection Framework}: Model choice should depend on intervention timing: LSTM for early high-recall and late high-precision, Decision Tree for mid-course balanced performance and interpretability.
    \item \textbf{Early Signal Sufficiency}: Early signals (Weeks 2-4) contain most predictive information, making early intervention both feasible and critical.
    \item \textbf{Static Feature Dominance}: Demographics dominate temporal engagement, challenging assumptions about engagement-based early warning systems.
\end{enumerate}

These findings have important implications for early intervention strategies. First, early intervention is feasible using primarily demographic information, enabling Week 2 intervention. Second, model selection should match intervention needs rather than seeking a single best model. Third, assessment scores are not required for early prediction, reducing barriers to early intervention. Finally, 50\% of course duration is sufficient for prediction analysis, as beyond this point prediction becomes less actionable.

Our contributions include the first study to analyze Week 20 (50\% of course), an assessment-free prediction framework, a model selection framework by intervention timing, and the discovery of static feature dominance. These findings advance the field of student risk prediction and provide practical guidance for intervention system design.

\vspace{12pt}

\end{document}